\documentclass{article}
\usepackage{spconf,amsmath,graphicx}
\usepackage{amssymb}
\usepackage{animate}

\DeclareMathOperator*{\argmin}{arg\,min}

\title{Photorealistic Image Reconstruction from Hybrid Intensity and Event based Sensor}
\name{Prasan Shedligeri \hspace{5pt} Kaushik Mitra}
\address{Indian Institute of Technology Madras, India\\
{\tt\small \{ee16d409, kmitra\}@ee.iitm.ac.in}}
\begin{document}
%
\maketitle
\begin{abstract}
Event sensors output a stream of asynchronous brightness changes (called ``events'') at a very high temporal rate. Previous works on recovering the lost intensity information from the event sensor data have heavily relied on the event stream, which makes the reconstructed images non-photorealistic and also susceptible to noise in the event stream.
We propose to reconstruct photorealistic intensity images from a hybrid sensor consisting of a low frame rate conventional camera, which has the scene texture information, along with the event sensor.
To accomplish our task, we warp the low frame rate intensity images to temporally dense locations of the event data by estimating a spatially dense scene depth and temporally dense sensor ego-motion. The results obtained from our algorithm are  more photorealistic compared to any of the previous state-of-the-art algorithms. We also demonstrate our algorithm's robustness to abrupt camera motion and noise in the event sensor data.
\end{abstract}

\section{Introduction}

\label{sec:intro}
Event-based sensors \cite{lichtsteiner2008128} encode the local contrast changes in the scene as positive or negative events at the instant they occur. Event-based sensors provide a power efficient way of converting the megabytes of per-pixel intensity data into a stream of spatially sparse but temporally dense events. However, the event stream cannot be directly visualized like a normal video, with which we as human beings are familiar with. 
This calls for an algorithm that can convert this stream of event data to a more familiar version of image frames. These reconstructed intensity frames could also be used as an input for traditional frame-based computer vision algorithms like multi-view stereo, object detection etc. Previous attempts \cite{reinbacher_bmvc2016,bardow2016simultaneous,barua2016direct, Scheerlinck18accv} at converting the event stream into images have heavily relied on event data. Although these methods do a good job of recovering the intensity frames they suffer from two major disadvantages: a) The intensity frames don't look photorealistic and b) some of the objects in the scene can go missing in the recovered frames because they are not producing any events (edges parallel to the sensor motion do not trigger any events).
\begin{figure}[t!]
    \centering
    \includegraphics[width=0.43\textwidth]{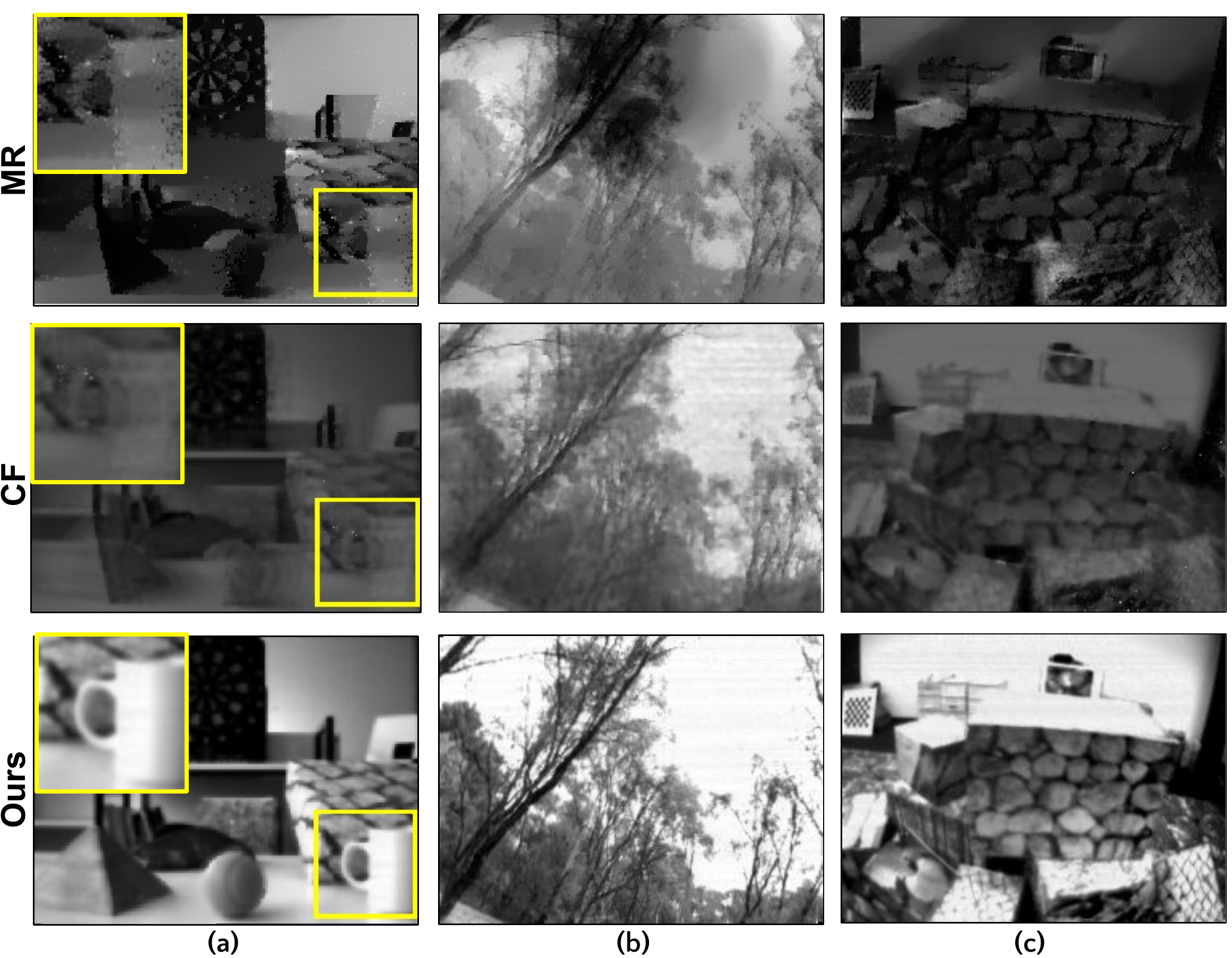}
    \vspace{-5pt}
    \caption{We compare our reconstruction from a hybrid sensor data (such as DAVIS) with that of CF \cite{Scheerlinck18accv} and MR\cite{reinbacher_bmvc2016}. Note that, MR only uses events for reconstruction. In column (a) we show in inset the zoomed-in version of an image region. We can clearly see that our proposed reconstruction method is able to recover the image region well compared to other state-of-the-art methods.
    }
    \label{fig:Photorealistic}
    \vspace{-10pt}
\end{figure}

\begin{figure}[t!]

\centering
\includegraphics[width=0.48\textwidth]{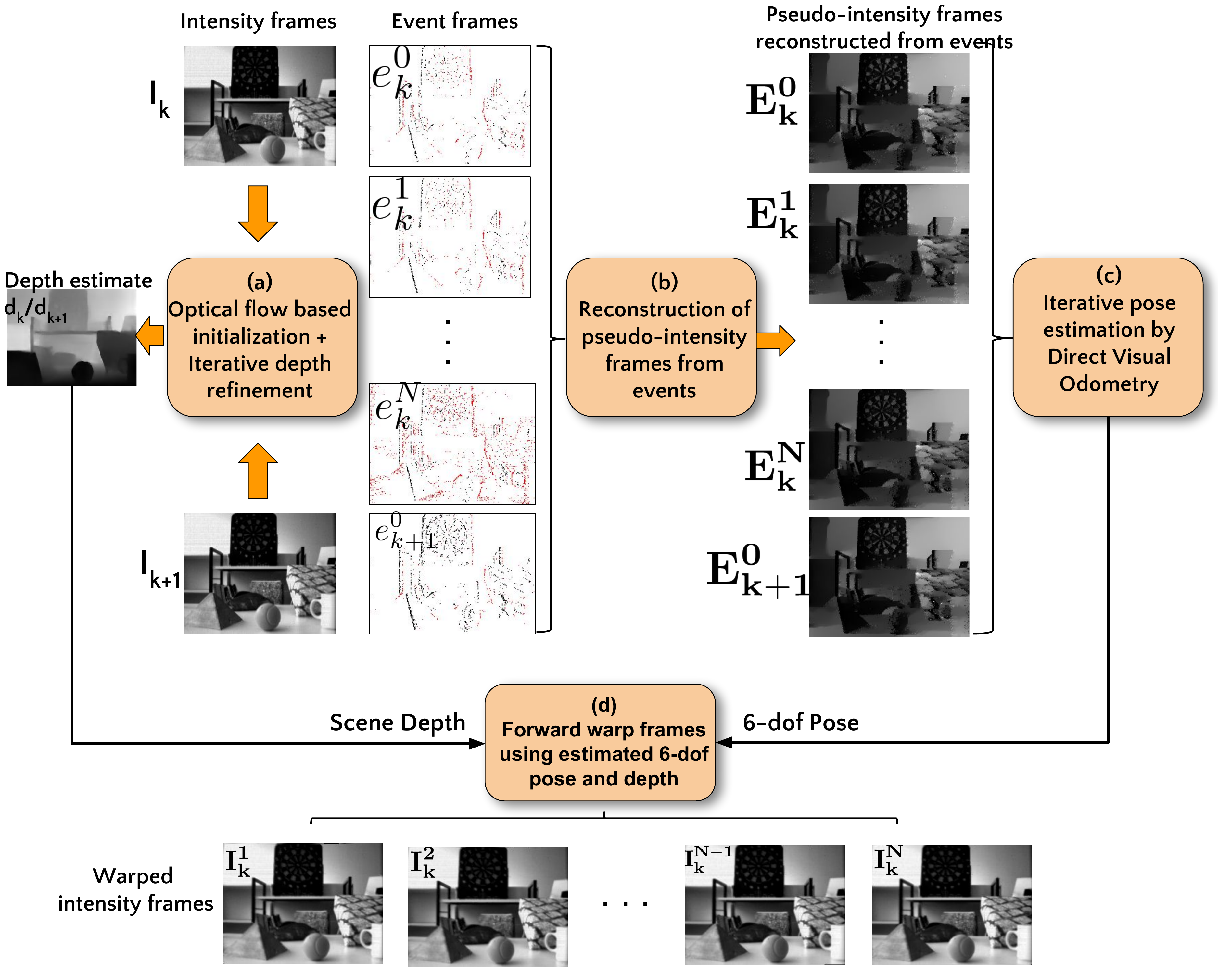}
\caption{Overview of our approach: The main blocks of our algorithms are a) an iterative depth and camera pose estimation technique for successive intensity frames, b) mapping event data into pseudo-intensity frames using \cite{reinbacher_bmvc2016}, c) direct visual odometry based sensor ego-motion estimation for intermediate event frame locations and d) a warping module for warping intensity images to intermediate locations.
}
\label{fig:teaser}
\end{figure}

In this paper, we propose a method to reconstruct photorealistic intensity images at a high frame rate. As the absolute intensity and fine texture information is lost during the encoding of events, we use the conventional image sensor to provide us with this information for reconstructing photorealistic images. There exists a commercially available hybrid sensor consisting of a co-located low-frame rate intensity sensor and an event-based sensor called DAVIS(Dynamic and Active pixel Vision Sensor)\cite{berner2013240}. 
Fig.~\ref{fig:teaser} summarizes our overall approach to reconstruct the temporally dense photorealistic intensity images using the hybrid sensor. We mainly have four steps. In the first step, we estimate the dense depth map from successive intensity frames. For this purpose, we use a traditional iterative optimization scheme, which we initialize by depth map obtained from a deep learning based optical flow estimation algorithm. In the second step, we map the event data between successive intensity frames to multiple pseudo-intensity frames using \cite{reinbacher_bmvc2016}. Next, we use the pseudo-intensity frames and the dense depth maps obtained from the first step to estimate temporally dense camera ego-motion by direct visual odometry. And finally, in the fourth step, we warp the successive intensity frames to intermediate temporal locations of the pseudo-intensity frames to obtain photo-realistic reconstruction. Recently, \cite{Scheerlinck18accv} have shown that it is possible to fuse temporally dense events with low frame-rate intensity frames to reconstruct intensity frames at a higher frame rate. However, due to lack of any regularization, the intensity frames reconstructed using \cite{Scheerlinck18accv} tend to be noisy and blurry. With extensive experiments, we show that our proposed method is able to reconstruct photorealistic intensity images at a high frame rate and is also robust to noisy events in the event stream. To summarize, we make the following contributions:
\begin{itemize}
    \item We propose a pipeline using a hybrid event and low frame rate intensity sensor which can reconstruct {\emph{temporally dense photorealistic intensity images}}. This would be difficult to obtain with only either the conventional image sensor or the event sensor.
    \item  We use the event sensor for estimating temporally dense sensor ego-motion and the low-frame rate intensity images for obtaining spatially dense depth map.
    \item We demonstrate high quality temporally dense photorealistic reconstructions using the proposed method on real data captured from DAVIS.
    \item We also demonstrate our algorithm's robustness to abrupt camera motion and noisy events in the event sensor data.
\end{itemize}

\section{Related work}
\textbf{Intensity image reconstruction from events:} The proposed work is very closely related to other previous works which reconstruct intensity images from events \cite{Scheerlinck18accv,reinbacher_bmvc2016,bardow2016simultaneous,barua2016direct}. \cite{reinbacher_bmvc2016,bardow2016simultaneous,barua2016direct} cannot recover the true intensity information of the scene as they use only the events to estimate the intensity images. Some works like \cite{kim2016real,rebecq2017evo} reconstruct intensity images as a by-product of sensor tracking from event data over 3D scenes but are not able to recover the true intensity information.
Recently, \cite{Scheerlinck18accv} demonstrated that event data and the intensity image data can be used in a complementary filter to reconstruct intensity frames at a higher frame rate. Although \cite{Scheerlinck18accv} makes use of the intensity images, the reconstructed images tend to be blurry and are adversely affected by noisy events due to lack of any regularization in their proposed method. The pre-print version of this work is also available on arxiv \cite{shedligeri2018photorealistic}.

\noindent \textbf{Visual odometry/SLAM with event sensors:} The high temporal data acquisition of event sensors has made it extremely suitable for applications such as tracking which need low latency operation. Very recently, a dataset was also proposed to benchmark event based pose estimation, visual odometry, and SLAM algorithms \cite{mueggler2017event}. The dataset contains multiple video sequences captured with DAVIS and sub-millimeter accurate ground truth camera motion acquired using a motion-capture system. Previous works such as \cite{vasco2017independent,reinbacher2017real,gallego2017event,reverter2016neuromorphic} estimate ego-motion of the sensor directly from the event stream. Visual odometry/ SLAM with event sensors \cite{rebecq2017evo,kim2016real,kueng2016low} has also been a very popular topic of research. Although we estimate the scene depth and sensor ego-motion to warp intensity frames, visual odometry/SLAM is not the focus of this work.

\section{Photorealistic image reconstruction}
We propose to reconstruct photorealistic intensity images using the event stream obtained from an event sensor. 
The conventional image sensor will compensate for the fine texture and the absolute intensity information which is lost in the event stream. As can be seen from Fig.~\ref{fig:teaser}, we have four major steps to reconstruct the temporally dense photorealistic image reconstruction: (a) Estimate dense depth maps $d_k$ and $d_{k+1}$ corresponding to the successive intensity frames $I_k$ and $I_{k+1}$ and the relative pose $\xi$ between them (\S \ref{sec:depthEst});
(b) Reconstruct pseudo-intensity frames $E_k^j$ at uniformly spaced temporally dense locations $j=1,2,\ldots N$ between every successive intensity frame $I_k$ and $I_{k+1}$; (c) Estimate temporally dense sensor ego-motion estimates $\xi_k^j$ and $\xi_{k+1}^j$ for each intermediate pseudo-intensity frame with respect to the intensity frames $I_k$ and $I_{k+1}$(\S \ref{sec:poseEst}) and (d)Forward warp the intensity frames $I_k$ and $I_{k+1}$ to the intermediate location of each of the pseudo-intensity frames $E_k^j$ and blend them (\S \ref{sec:warping}).
\subsection{Depth estimation from two successive intensity images}
\label{sec:depthEst}

\begin{figure}[t!]
\centering
\includegraphics[width=0.48\textwidth]{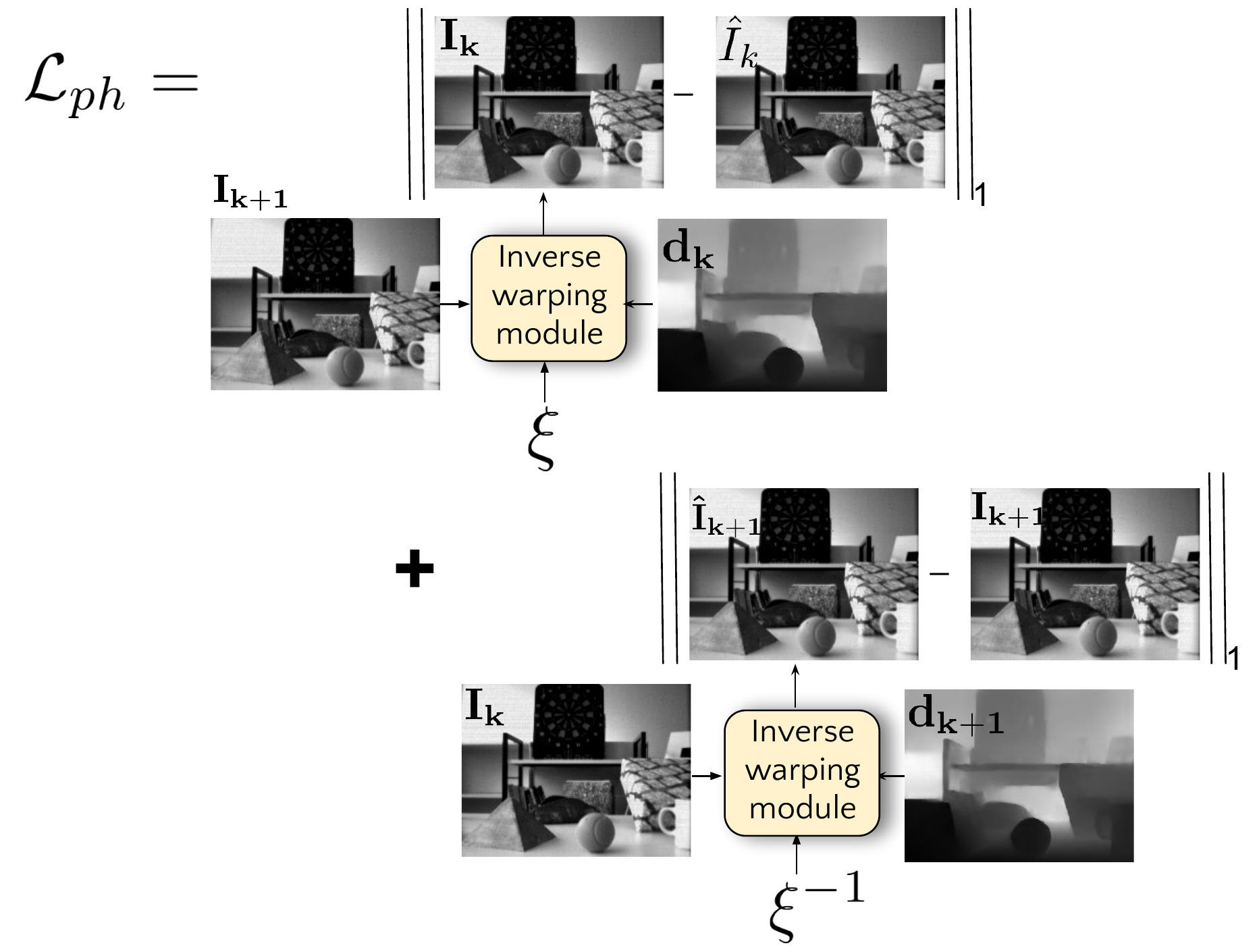}
\caption{
\emph{Estimating dense depth maps and relative pose of two successive intensity images:} 
We use the optical flow estimated from PWC-Net \cite{Sun2018PWC-Net} to obtain an initial depth estimate and initialize the relative pose to zero rotation and translation. We iteratively minimize the photometric error over the depth maps $d_k$ and $d_{k+1}$ and the relative pose $\xi$.
}
\label{fig:depthEst}

\includegraphics[width=0.48\textwidth]{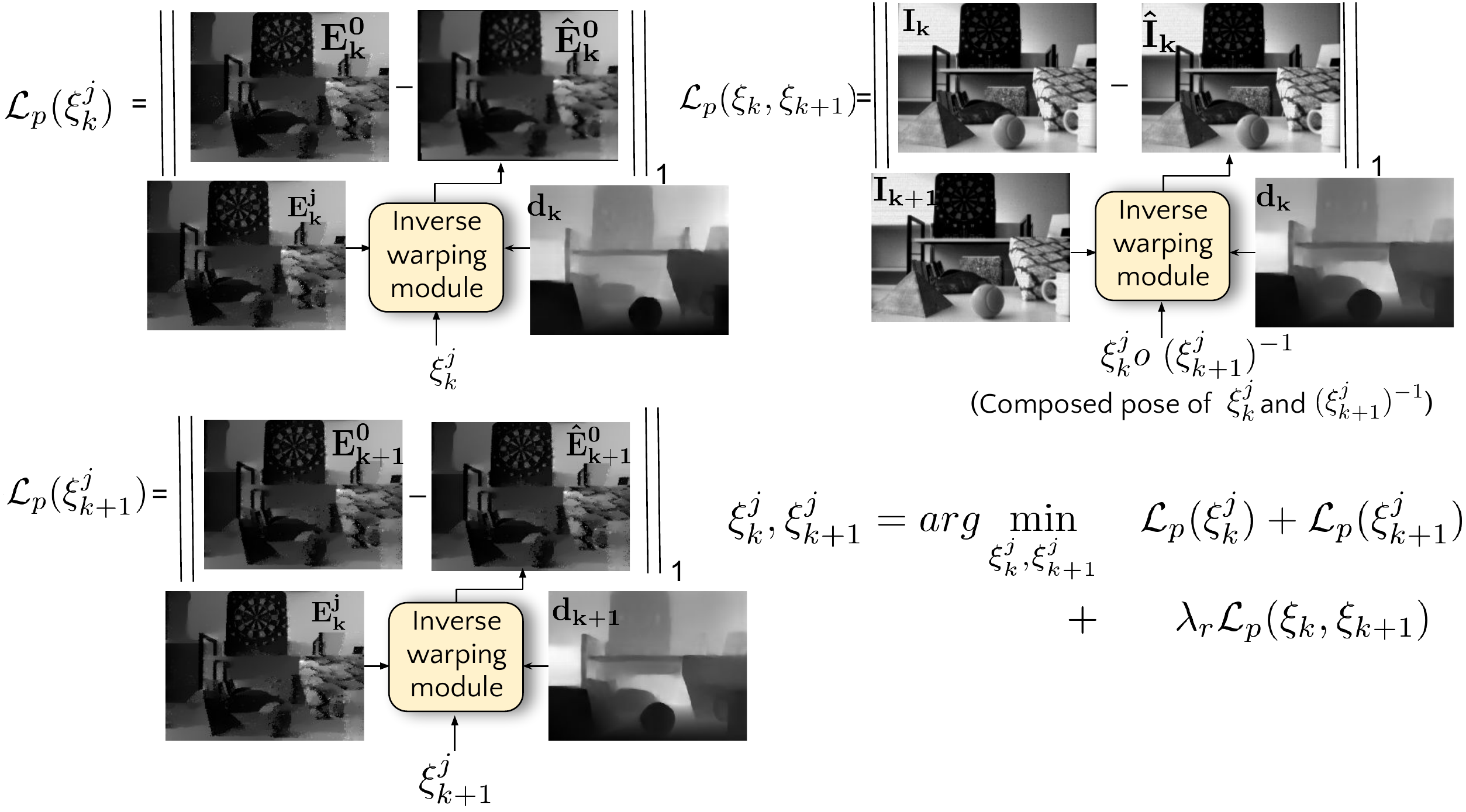}
\caption{
\emph{Estimating relative pose of intermediate pseudo-intensity images:} 
We estimate the 6-dof camera pose of $E_k^j$ w.r.t. $E_k^0$ and $E_{k+1}^0$ by iteratively minimizing the photometric error between the warped image $\hat E_k^j$ and the target image $E_k^j$. We minimize the photometric error over the relative poses $\xi_k^j$ and $\xi_{k+1}^j$ using the known depth estimates $d_k$ and $d_{k+1}$.
}
\label{fig:poseEst}
\end{figure}

One of the important steps in our proposed algorithm is forward warping the intensity images to multiple intermediate temporal locations between successive intensity frames. However, warping can introduce undesired holes in the final reconstructed images at regions of disocclusion. This can be solved by warping both the successive intensity frames, $I_k$ and $I_{k+1}$, to the intermediate locations. This requires us to estimate two dense depth maps $d_k$ and $d_{k+1}$ corresponding to the images $I_k$ and $I_{k+1}$, respectively.
Fig. \ref{fig:depthEst} shows the overall scheme of estimating dense depth maps from successive intensity frames. We initialize the depth estimates $d_k$ and $d_{k+1}$ from optical flow, and the 6-dof camera pose $\xi$ with zero rotation and translation. Here, $\xi$ is the 6-dof relative camera pose at $I_{k+1}$ with respect to $I_k$. We warp the intensity image $I_{k+1}$ to the location of $I_{k}$ with the current estimate of $d_k$ and $\xi$, to obtain $\hat I_{k}$. Similarly, we warp $I_k$ to the location of $I_{k+1}$ to obtain $\hat I_{k+1}$.
We define the photometric reconstruction loss $\mathcal{L}_{ph}$ as,
\begin{equation}
\mathcal{L}_{ph}(d_k,d_{k+1},\xi) = \|(\hat{I}_k - I_k)\|_1 + \|(\hat{I}_{k+1} - I_{k+1})\|_1    
\end{equation}
By minimizing the above reconstruction loss, $\mathcal{L}_{ph}$, it is possible to estimate the depth maps $d_k$ and $d_{k+1}$ and 6-dof relative pose $\xi$. We also enforce an edge aware laplacian smoothness prior on the estimated depth maps $d_k$ and $d_{k+1}$, by taking inspiration from \cite{mahjourian2018unsupervised}.
We define the smoothness loss $\mathcal{L}_{sm}$ as,
\begin{equation}
    \mathcal{L}_{sm}(d) = \sum \|\nabla_x d\| e^{- \beta \|\nabla_x I\|} + \|\nabla_y d\| e^{- \beta \|\nabla_y I\| }
    \label{eq:DepthSmooth}
\end{equation}
where $I$ is the intensity image, $d$ is the corresponding dense depth map and $\nabla_x$ and $\nabla_y$ are the x and y-gradient operators, respectively. Overall, we estimate the dense depth estimate $d_k$, $d_{k+1}$ and the relative pose $\xi$ by,
\begin{multline}
    \label{eq:depthEst}
    \xi,d_k,d_{k+1} = \argmin_{\xi,d_k,d_{k+1}} \mathcal{L}_{ph}(d_k,d_{k+1},\xi) \\+ \lambda_{sm}(\mathcal{L}_{sm}(d_k) + \mathcal{L}_{sm}(d_{k+1}))
    \vspace{-5pt}
\end{multline}

Eq.~\eqref{eq:depthEst} is a non-convex optimization problem and hence a good initialization of depth and pose is essential to avoid local minima. Here, we use optical flow between the successive intensity frames obtained from PWC-Net\cite{Sun2018PWC-Net} as an initial estimate of the depth.
For a static scene, it is possible to estimate the scene depth and the 6-dof camera pose from the optical flow. 
However, in our experiments, we found that a simple inverse of the optical flow magnitude is good enough to initialize the depth for the optimization iterations in Eq.~\eqref{eq:depthEst}. We initialize pose with zero rotation and translation.

\subsection{6-dof relative pose estimation by direct matching}
\label{sec:poseEst}
To achieve the goal of photorealistic reconstruction we warp the successive intensity frames captured by the image sensor to the intermediate temporal location of an event frame. For warping, we need to determine the 6-dof camera pose between the temporal locations of the successive intensity frames and that of the intermediate event frames. We reconstruct pseudo-intensity images from events using \cite{reinbacher_bmvc2016} at the temporal locations of the intermediate event frames as well as the successive intensity frames. 
As shown in Fig. \ref{fig:poseEst}, our goal here is to estimate the relative camera pose between $E_k^0$, $E_{k+1}^0$ and the pseudo-intensity images $E_k^j$ ($j= 1,2,\ldots N$). We use this relative pose, to warp the successive intensity frames to the intermediate locations specified by the event frames ($E_k^j$) and hence reconstruct photorealistic intensity images.

\noindent Let $\xi_k^j$ represent the 6-dof camera pose of the intermediate pseudo-intensity image $E^j_{k}$ with respect to $E^0_k$ and $\xi_{k+1}^j$ be the 6-dof camera pose of $E^j_{k}$ with respect to $E^0_{k+1}$. 
We use the current estimate of relative camera pose $\xi_{k}^j$ and the known depth estimate $d_k$ to inverse warp the pseudo-intensity frame $E^j_{k}$ to the location of $E^0_k$ to obtain $\hat E^0_k$. We similarly inverse warp the pseudo-intensity frame $E^j_{k}$ to the location of $E^0_{k+1}$ to obtain $\hat E^0_{k+1}$ using the current estimate of relative pose $\xi_{k+1}^j$ and the known depth $d_{k+1}$. We define the photometric loss $\mathcal{L}_{p}$ as mean absolute error between the warped intensity frame and the ground truth frame. 
\begin{align}
    \mathcal{L}_{p}(\xi_k^j) & = \| E^0_k - \hat{E}^0_k\|_1 \\
    \mathcal{L}_{p}(\xi_{k+1}^j)& = \| E^0_{k+1} - \hat{E}^0_{k+1}\|_1
\end{align}
By composing the relative pose estimates, $\xi_k^j$ and  $(\xi_{k+1}^j)^{-1}$ we obtain the overall pose between $I_k$ and $I_{k+1}$. We use this knowledge to regularize the relative camera pose estimates $\xi_k^j$ and  $\xi_{k+1}^j$ with $\mathcal{L}_{p}(\xi_{k}^j, \xi_{k+1}^j) = \| I_k - \hat{I}_k\|_1$ . Overall,
\begin{multline}
    \xi_{k}^j,\xi_{k+1}^j = \argmin_{\xi_{k}^j,\xi_{k+1}^j} \mathcal{L}_{p}(\xi_k^j) + \mathcal{L}_{p}(\xi_{k+1}^j) + \lambda_{r} \mathcal{L}_{p}(\xi_{k}^j, \xi_{k+1}^j)
    \label{eq:poseEst}
\end{multline}
where $\lambda_{r}$ is the regularization parameter.


\subsection{Forward Warping and Blending}
\label{sec:warping}
At this stage, we have depth maps $d_k$ and $d_{k+1}$ corresponding to intensity images $I_{k}$ and $I_{k+1}$ respectively. 
We do a source-target mapping (forward warping) from two images $I_k$ and $I_{k+1}$ using the estimated relative pose $\xi_k^j$ and $\xi_{k+1}^j$to the latent image $I_k^j$ and alpha-blend them. We splat the intensity values after forward warping to ensure that no holes are generated in the final image.

\section{Experiments}
For all our experiments we use DAVIS \cite{berner2013240}, which is commercially available and has a conventional image sensor and an event sensor bundled together. Since, we did not have access to DAVIS, we used the recently proposed dataset by \cite{mueggler2017event} and \cite{Scheerlinck18accv} which consists of several video sequences captured using DAVIS. We obtain dense depth maps at the locations of low frame rate intensity frames and temporally dense sensor ego-motion using the event sensor data to warp the low frame-rate intensity frames to intermediate camera locations. For estimating depth, we initially enhance the edges of the depth obtained from optical flow estimate using a fast bilateral solver \cite{barron2016fast}. The output of this bilateral solver is then used as an initialization for the iterative depth refinement scheme. We set $\beta = 10.0$ in Eq.~\eqref{eq:DepthSmooth} and $\lambda_{sm}=1.0$  in Eq.~\eqref{eq:depthEst}. Using the event stream from each sequence in the dataset we generate pseudo-intensity estimates using the algorithm proposed in \cite{reinbacher_bmvc2016}. We stack non-overlapping blocks of 2000 events into a frame and generate a corresponding pseudo-intensity frame using \cite{reinbacher_bmvc2016}. These pseudo-intensity frames are then used for estimating the temporally dense sensor ego-motion.
For pose estimation we use $\lambda_r=0.01$ in Eq.~\eqref{eq:poseEst}. We use the Adam optimizer \cite{kingma2014adam} to solve Eq.~\eqref{eq:depthEst} and Eq.~\eqref{eq:poseEst}.

\begin{figure}[b!]
\centering
\includegraphics[width=0.48\textwidth]{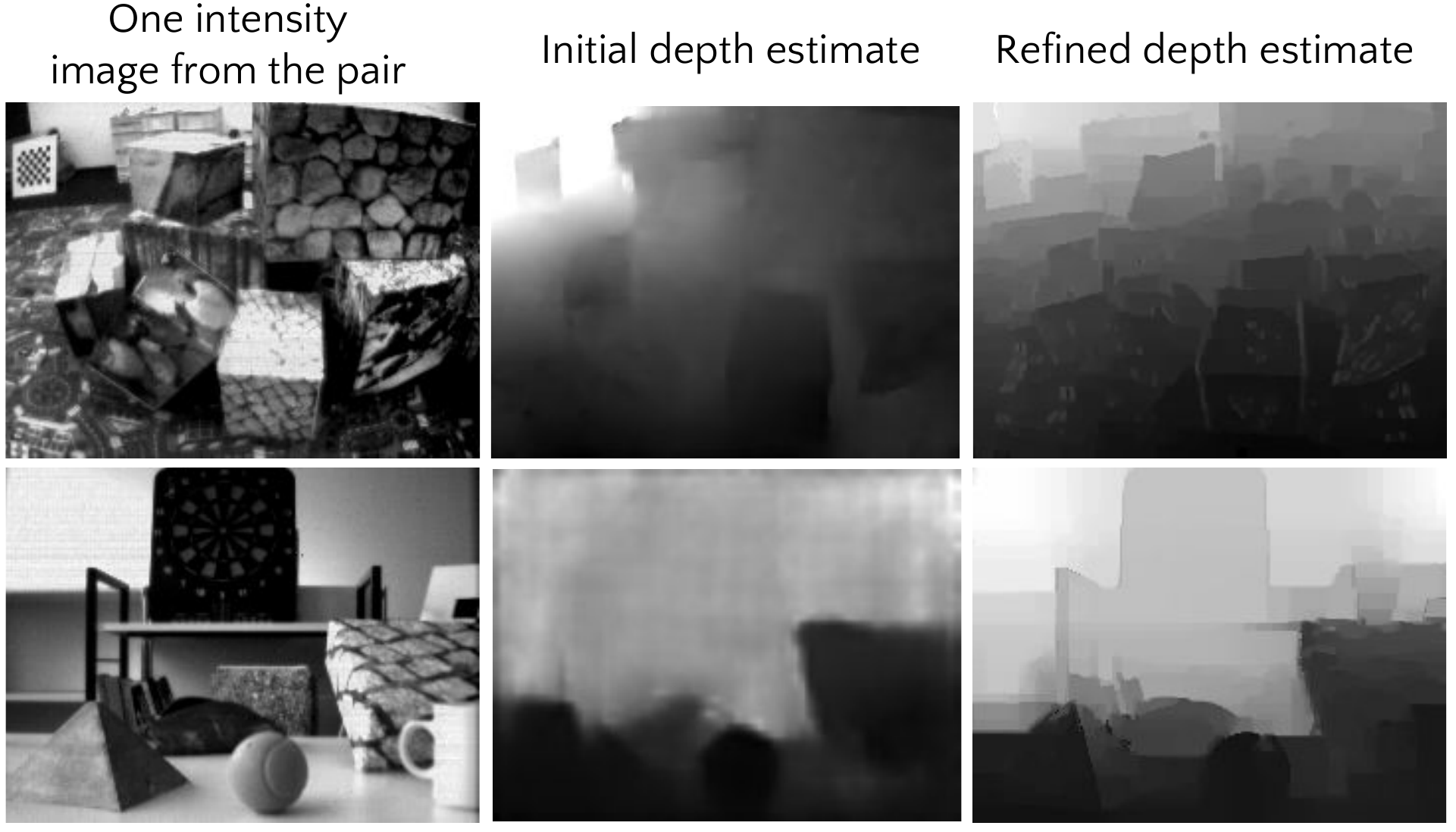}
\caption{Estimated depth maps from our proposed method}
\label{fig:depth}
\vspace{7pt}
\begin{tabular}{ccc}
Successive \\ intensity frames & Ours & CF\cite{Scheerlinck18accv} \\
\animategraphics[width=0.15\textwidth,loop]{2}{"images/seq6_seq/raw/frame_0000055"}{3}{4} \hspace{-8pt} & \animategraphics[width=0.15\textwidth,loop]{30}{"images/seq6_seq/ours/seq6-"}{1}{30} \hspace{-8pt} & \animategraphics[width=0.15\textwidth,loop]{10}{"images/seq6_seq/CF/seq6-"}{1}{10}\hspace{-8pt} \\
\animategraphics[width=0.15\textwidth,loop]{2}{"images/seq9_seq/raw/frame_0000055"}{3}{4} \hspace{-8pt} & \animategraphics[width=0.15\textwidth,loop]{12}{"images/seq9_seq/ours/seq9-"}{1}{12}\hspace{-8pt} & \animategraphics[width=0.15\textwidth,loop]{11}{"images/seq9_seq/CF/seq9-"}{1}{11}\hspace{-8pt} \\
\end{tabular}
\caption{Qualitative comparisons of reconstructions. \textit{(If document is opened in Adobe Reader, videos can be viewed by clicking on the images)}. We provide further video comparisons in the supplementary material.
}
\label{fig:more}
\vspace{5pt}
\includegraphics[width=0.48\textwidth]{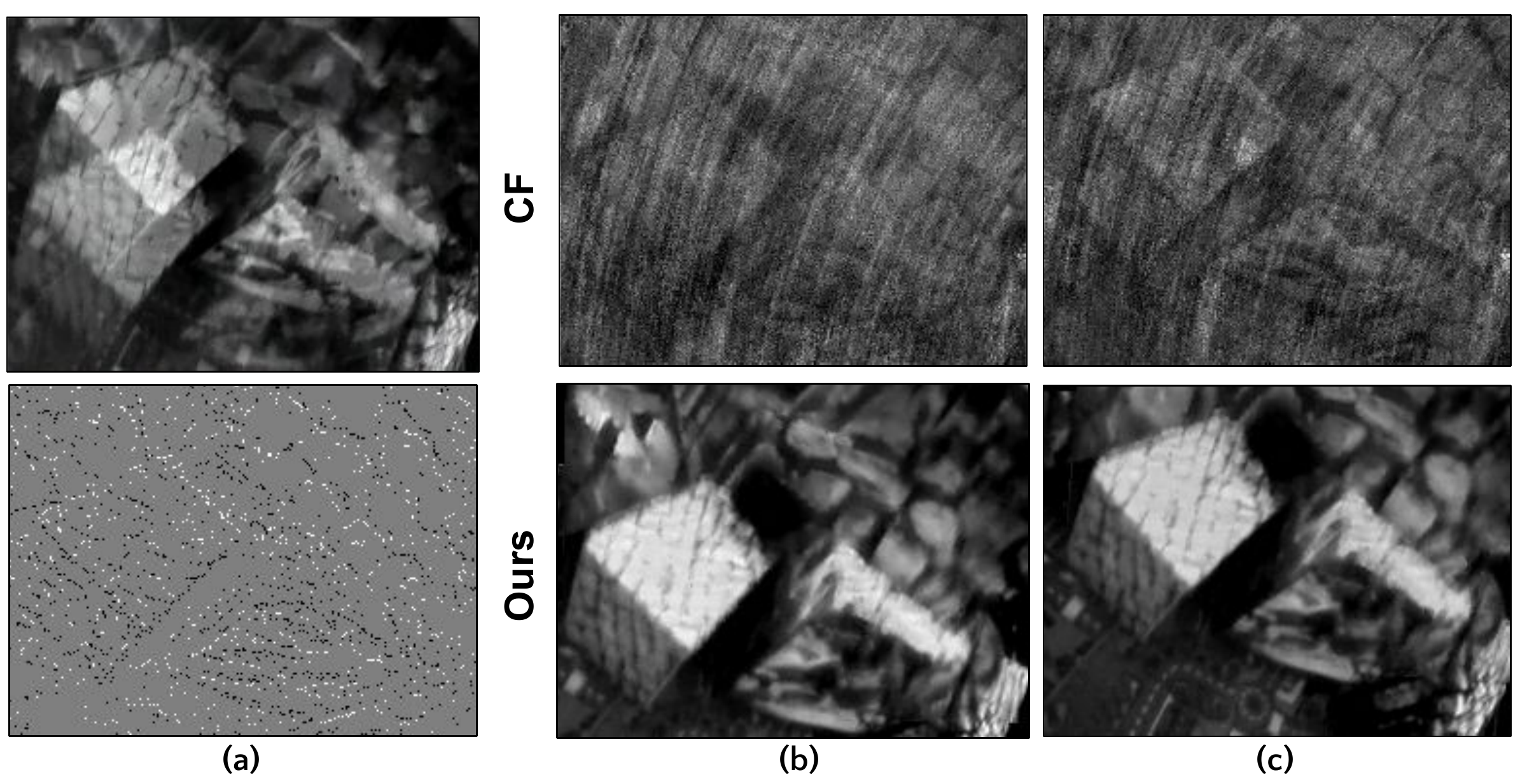}
\caption{\emph{Abrupt camera motion:}  (a)Top row shows the two successive images blended into one where we can see the abrupt camera motion. The bottom row shows an intermediate event frame affected by noise. (b),(c): Comparing reconstructions obtained from our method to that of CF \cite{Scheerlinck18accv} on the sequence with abrupt camera motion.
}
\label{fig:abrupt}
\end{figure}
\subsection{Depth estimation}
In Fig.~\ref{fig:depth} we demonstrate the effectiveness of our proposed method for estimating depth. We use an initial estimate of depth from a deep learning method and iteratively refine it. We empirically found that using PWC-Net \cite{Sun2018PWC-Net} to initialize the depth estimate for the iterative optimization scheme gave consistently good results. We also experiment with initializing the depth from \cite{ummenhofer2017demon,IMKDB17_flownet}. We provide comparisons in supplementary material.
\subsection{Photorealistic intensity image reconstruction}
In Fig.~\ref{fig:Photorealistic} and Fig.~\ref{fig:more} we compare qualitatively the intensity images reconstructed using our proposed method to that proposed in \cite{reinbacher_bmvc2016,Scheerlinck18accv}. While MR\cite{reinbacher_bmvc2016} utilizes only event sensor data, CF\cite{Scheerlinck18accv} uses both event sensor data as well as information from intensity images. For fairness in comparison, we generate intensity images from \cite{reinbacher_bmvc2016,Scheerlinck18accv} for every $2000$ events in the sequence. 
In \cite{Scheerlinck18accv}, we found that initializing the cut-off frequency to $6.28 rad/s$ and updating other parameters dynamically gave the best results.  

\noindent Reinbacher \textit{et al.} \cite{reinbacher_bmvc2016} use only event information and are hence unable to recover the true intensity information present in the scene. Scheerlinck \textit{et al.} \cite{Scheerlinck18accv} do not use any kind of spatial regularization and hence the reconstructed images are noisy and blurry even though they have access to the intensity images. We acknowledge that \cite{Scheerlinck18accv,reinbacher_bmvc2016} run in real time, while our algorithm takes about two minutes to estimate the dense depth maps and about 40 seconds to render each intermediate frame. With recent advances in stereo depth estimation methods, we expect that in future we can completely eliminate the need for an iterative depth refinement scheme and directly use the output of a state-of-the-art stereo depth estimation algorithm. This will greatly reduce the computation time. It is possible to further reduce the computation time for estimating pose by using the Lucas-Kanade inverse compositional method.

\subsection{Robustness to abrupt camera motion}
In the case of abrupt motion of the sensor, the intensity images get blurred and the rate at which events are generated becomes high. We start with deblurring the intensity images using an existing deblurring technique(in our experiments we used \cite{Nah_2017_CVPR}). These deblurred images are then used as an input to the reconstruction pipeline. Abrupt motion results in a high event rate and also produces many noisy events. These noisy events affect the reconstructions in \cite{Scheerlinck18accv} as their trust on events increases exponentially with the rise in the event rate. As can be seen in Fig.~\ref{fig:abrupt}, our method is robust to such abrupt motions as can be seen from the results shown in columns (b) and (c).
\section{Conclusion}
We combine the strength of texture-rich low frame rate intensity frames with  high temporal rate event data to obtain temporally dense photo-realistic images. We achieve this by warping the low frame rate intensity frames from the conventional image sensor to intermediate locations. 
With extensive experiments, we have demonstrated that the images reconstructed from our algorithm are photorealistic compared to any of the previous methods. We also show the robustness of our algorithm to abrupt camera motion. 
Currently, our algorithm assumes a static scene. A future direction for us would be to build a generalized algorithm which can reconstruct photorealistic images for dynamic scenes as well. 


\end{document}